\documentclass[letterpaper]{article} 
\usepackage{aaai23}  
\usepackage{times}  
\usepackage{helvet}  
\usepackage{courier}  
\usepackage[hyphens]{url}  
\usepackage{graphicx} 
\usepackage{natbib}  
\usepackage{caption} 
\usepackage{algorithm}
\usepackage{algorithmic}
\usepackage{bm}
\usepackage{makecell}
\newcommand{\specialcell}[2][c]{%
  \begin{tabular}[#1]{@{}l@{}}#2\end{tabular}}
\usepackage{newfloat}
\usepackage{listings}
\DeclareCaptionStyle{ruled}{labelfont=normalfont,labelsep=colon,strut=off} 
\floatstyle{ruled}
\newfloat{listing}{tb}{lst}{}
\floatname{listing}{Listing}
\title{A Data Fusion Framework for Multi-Domain Morality Learning}
\author{
    Siyi Guo\textsuperscript{\rm 1}, Negar Mokhberian\textsuperscript{\rm 1}, Kristina Lerman\textsuperscript{\rm 1}
}
\affiliations{
    \textsuperscript{\rm 1}Information Sciences Institute, University of Southern California\\ 
    Marina del Rey, CA, USA\\
    \{siyiguo,nmokhber,lerman\}@isi.edu

}

\usepackage{bibentry}

\begin{document}

\maketitle

\begin{abstract}
Language models can be trained to recognize the moral sentiment of text, creating new opportunities to study the role of morality in human life. As interest in language and morality has grown, several ground truth datasets with moral annotations have been released. However, these  datasets vary in the method of data collection, domain, topics, instructions for annotators, etc. Simply aggregating such heterogeneous datasets during training can yield models that fail to generalize well. We describe a data fusion framework for training on multiple heterogeneous datasets that improve performance and generalizability. The model uses domain adversarial training to align the datasets in feature space and a weighted loss function to deal with label shift. We show that the proposed framework achieves state-of-the-art performance in different datasets compared to prior works in morality inference.
\end{abstract}

\section{Introduction}

Morality helps people distinguish between ``right'' and ``wrong'' and governs their everyday behaviors and interactions with others~\cite{hofmann2014morality}. Morality also shapes judgments, attitudes and beliefs, creating differences in the moral experience of individuals across cultural groups~\cite{haidt2007moral}. Studies have linked moral sentiment to partisan ideologies~\cite{graham2009liberals}, messaging strategies in politics~\cite{wang2021moral} and news~\cite{mokhberian2020moral}, and even real-world violence~\cite{mooijman2018moralization}.

Researchers have developed a scale to quantify morality, which represents people's intuitive ethical reactions to social dilemmas. The so-called Moral Foundations Theory (MFT)~\cite{Haidt2004-HAIIEH,GRAHAM201355}  characterizes morality along five dimensions:
\begin{itemize}
\item \textit{Care/Harm}: dislike of suffering.
\item \textit{Fairness/Cheating}: proportionality, justice and rights.
\item \textit{Loyalty/Betrayal}: attachment to one's identified group.
\item \textit{Authority/Subversion}: respect for authority and tradition.
\item \textit{Sanctity/Degradation}: endorsement of purity and cleanliness, avoidance of pollution and decay.
\end{itemize}
While early research in morality science relied on moral foundation questionnaires and vignettes to characterize morality along these dimensions~\cite{graham2011mapping}, recent works have begun to infer morality from text. Automated natural language processing (NLP) methods \cite{fulgoni-etal-2016-empirical,8b2871f503a14011ae81e6ab1664a638,lin2018acquiring,xie2020contextualized} have enabled researchers to scale up moral foundation inference to large text corpora of news articles and messages posted on social media, opening new avenues for studying morality. 

The more sophisticated approaches train language models on ground truth datasets of text labeled by human annotators according to its moral expression. The trained models are then used to recognize morality of a new text. The rapid growth of interest in language and morality has produced several ground truth datasets for moral foundation inference \cite{Hoover2020moral,johnson-goldwasser-2018-classification,forbes-etal-2020-social,hopp2021extended,trager2022moral}.  
Researchers hope that training models on multiple datasets will yield better performance and generalizability. However, the labeled datasets are \textit{heterogeneous}: they vary by domain (news vs social media), topics covered (politics vs health), task granularity (label the moral foundations vs its vices and virtues), annotator population (few experts vs a crowd), annotation instructions (select text expressing a given moral foundation vs select relevant moral foundations for a given text), etc. Blindly combining heterogeneous data during training may bias predictions~\cite{bareinboim2016causal}. For example, models trained on aggregated data may give predictions that are inconsistent with predictions of models trained on disaggregated data, a phenomenon called Simpson's paradox~\cite{alipourfard2021disaggregation}.

We address the challenge of training moral foundation classifiers on heterogeneous datasets via a data fusion-inspired approach. 
Unlike training a classifier on aggregated data, our proposed approach uses domain adversarial training \cite{ganin2015unsupervised} to map the features (text embeddings) from different datasets onto a common embedding space, thus mitigating the problem brought by the heterogeneity of the texts and topics in the datasets and improves the generalizability across different data sources.

In addition to differences in the feature space, datasets can also vary in the distribution of labels. For example, tweets related to the pandemic have many more messages expressing the \textit{care/harm} foundation than tweets related to civil unrest. Unless accounted for, differences in label distribution will hurt classifier performance. To mitigate this problem, we propose to use a weighted loss function that balances among different label classes and between positive and negative data examples. 

Compared to previously used morality detection methods, our proposed framework achieves state-of-the-art performance on many datasets in out-of-domain testing. We believe that our work is the first one improving the generalizability of models for moral foundation inference with multi-dataset learning and a domain adaptation approach.

\section{Related Works}

\paragraph{Morality Inference}

Many prior works have contributed to the development of methods to classify moral foundations from text. One type of methods is dictionary-based, which relies on the use of lexical resources, such as the Moral Foundations Dictionary (MFD) \cite{Haidt2004-HAIIEH,haidt2007morality}. 
Some researchers proposed to use distributed dictionary representations (DDR) to capture the semantic similarities between words \cite{8b2871f503a14011ae81e6ab1664a638}. Another method measured the distance of a text from the axes defined by the words representing the virtues and vices of the moral foundations in the embedding space~\cite{mokhberian2020moral}. 

With recent advances in transformer-based language models such as BERT \cite{devlin2019bert}, researchers have shown that these large pre-trained language models have a sense for social norms. \citet{schramowski2022large} demonstrated that BERT captures general morality and identifies right or wrong actions. Some studies have leveraged these language models to calculate contextualized embeddings for moral foundations inference \cite{xie2020contextualized,PPR:PPR321781, hofmann-etal-2022-modeling}. Other researchers have made effort in gathering and annotating datasets for moral foundations analysis  \cite{Hoover2020moral,trager2022moral}.

As the interest in studying morality grows, especially on large quantities of social media data, inferring morality on large unlabeled data becomes a common challenge. Thus, how to utilize limited resources and perform out-of-domain inference becomes an important research question. \citet{islam2022understanding} proposed a minimally supervised framework by learning on a combination of many weak labels and a smaller amount of gold labels to analyze morality related to the COVID vaccine topic. \citet{pacheco-etal-2022-holistic} also studied morality detection in a low-resource setting to analyze COVID-19 vaccine debates. Their model is trained on a larger quantity of available out-of-domain labeled data and finetuned on a small amount of in-domain labels. 

However, in-domain gold labels for targeted data are not always available. Annotating morality is an especially challenging task, suffering from great time complexity and low annotator agreement. In this work, we explore a more challenging setting, performing out-of-domain morality inference without any available in-domain labels. We propose to utilize already available heterogeneous datasets with morality labels, and adopt domain adaptation ideas to improve model's out-of-domain performance. 

\paragraph{Domain Adaptation}

The unsupervised domain adaptation problem is an active research area in machine learning. Researchers have developed various approaches, including structural correspondence learning \cite{blitzer-etal-2006-domain}, joint distribution matching using max mean discrepancy \cite{6751384} and mixture of experts \cite{guo-etal-2018-multi}. 
Other recent works have tried to improve out-of-domain prediction by improving the training data quality. \citet{le2020adversarial} studied AFLite, a method that cleans the redundant information in training data thus mitigates model's dependence on spurious correlations.

Another popular method is the domain adversarial neural network (DANN) \cite{ganin2015unsupervised}. On top of a regular feature extractor and a label classifier, this model structure includes a domain classifier, an adversary that pushes the feature extractor to generate domain-invariant features, and thus facilitates domain adaptation. This architecture is shown to be successful in other NLP tasks such as aspect-dependent text classification \cite{zhang2017aspect} and stance detection \cite{allaway2021adversarial,hardalov2021cross}. DANN is a great method to deal with feature distribution shift, which is one of the major problems in merging heterogeneous datasets with different topics and annotation processes. Thus, we incorporate the DANN structure as a part of our model.

The similar idea of merging multiple datasets and incorporating domain adaptation techniques has been tested out in prior works for stance detection \cite{hardalov2021cross,li-etal-2021-improving-stance}. However, stance detection is very different from morality detection, as their main challenges are the varying label sets (e.g. for/against in one dataset, comment/support/query/deny in another dataset) and varying target sets (e.g. Trump, climate change, legalization of abortion, etc). These prior works have thus put more focus on adapting among different label and target sets. Morality prediction, on the other hand, is backed with the well-defined Moral Foundations Theory and has the same set of labels for most of our datasets. Instead, our major challenge is the feature and label distribution shift among datasets. On top of merging datasets and using domain adversarial training, we propose to use a weighted loss function to balance between different label classes.



\section{Methods}

\begin{figure*}[htb!]
\centering
\includegraphics[width=0.8\textwidth]{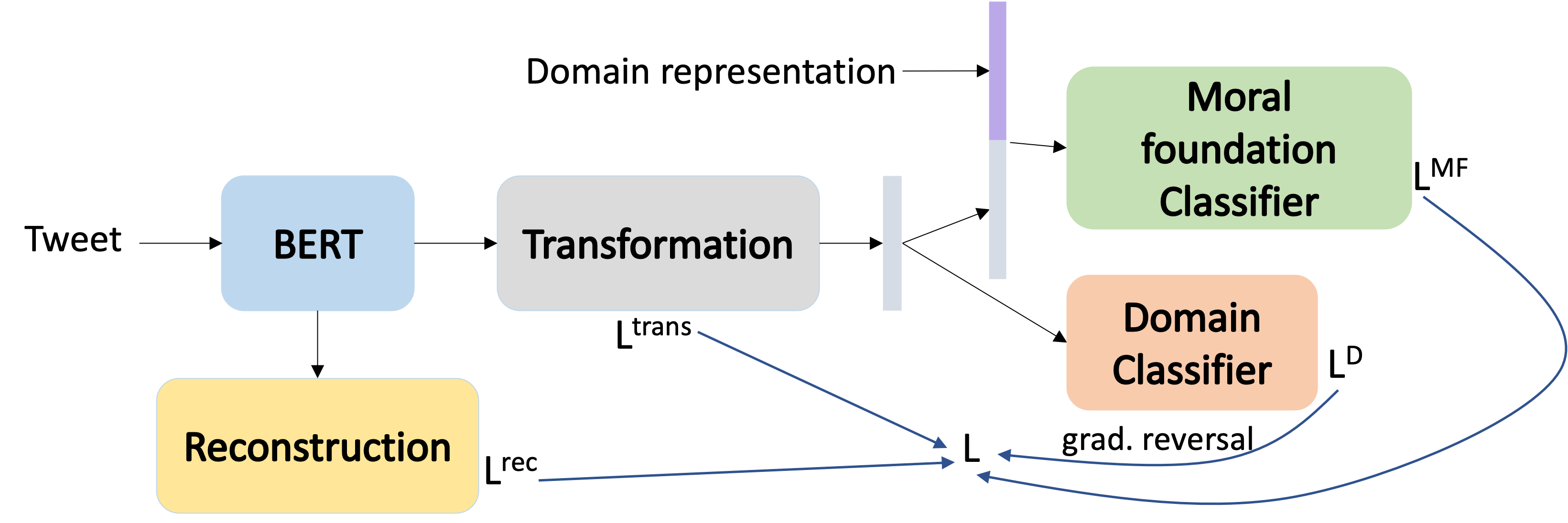}
\caption{Model structure of \textbf{DAMF}: BERT working as text encoder, a transformation module facilitating the text embedding to be more domain-invariant, a moral foundation classifier with a weighted loss function, an adversarial domain classifier, and a reconstruction module to avoid the model being corrupted by adversarial training. The objective is to learn domain-invariant embeddings that can fool the domain classifier and also performs well in morality detection.} 
\label{fig_model}
\end{figure*}

We propose a Domain Adapting Moral Foundation inference model (\textbf{DAMF}) for fusing annotated data from multiple domains. The model (Figure \ref{fig_model}) has five parts. BERT encodes texts into contextualized embeddings. The transformation module facilitates the text embeddings to be more domain-invariant. The moral foundation classifier with a weighted loss function performs the main morality detection task. The domain classifier works as the adversary, pushing the BERT encoder to learn domain-invariant embeddings to fool the domain classifier. Additionally, the reconstruction module attached to the BERT encoder avoids the model from being corrupted by adversarial training. The objective of the model is to learn domain-invariant embeddings to allow heterogeneous domains to align in the embedding space, and also to succeed in morality detection. 


\paragraph{Text Encoder}

We use the pre-trained, transformer-based language model BERT to transform input messages into embedding vectors.

\paragraph{Domain-invariant Transformation}

The output embeddings from the text encoder contains domain-specific information. To facilitate the transfer across different domains, we follow the idea from Zhang et al. (\citeyear{zhang2017aspect}) and add a linear domain-invariant transformation layer:
\[x^{trans} = W^{trans}x^{BERT}\]

\noindent
where $W^{trans}$ is the weight for the linear layer, $x^{BERT}$ is the embedding generated from the BERT text encoder, and $x^{trans}$ is the embedding after this transformation layer. This transformation is regularized to the identity \(I\) to ensure text information is still retained and won't be wiped out by the adversary. Therefore, from this step we obtain a loss term \(L^{trans}\):
\[L^{trans} = || W^{trans} - I ||^2\]

\paragraph{Moral Foundation Classifier}

This module is in charge of learning the primary task: classifying moral foundations. It has a linear layer, a ReLu activation, a dropout layer, and a second linear layer. The input to this classifier is the embeddings generated by the previous steps (\(x^{trans}\)) concatenated to a one-hot domain embedding. This is to facilitate the model learning the relationship between the input tweet and its domain. 

One tweet can be related to multiple moral foundations. For example, this tweet during the COVID-19 outbreak ``I'm tired of people and media talking about all the merchandise that been looted, what about the minority lives that been lost'' expresses both harm and cheating moral foundations. Therefore, we formulate moral foundation inference as a multi-label classification task. There are 10 classes in total (e.g. care, harm, etc) and each class itself is a binary prediction. 

\paragraph{Weighted Loss Function}

One problem in moral foundation inference is that the number of positive and negative examples is severely imbalanced. That is, we have many more non-moral data than those with morality labels. Moreover, the number of examples associated with different moral foundation classes can also vary greatly (Figure \ref{fig_label_distr}). 
To balance between positive and negative examples, as well as among different classes, we use a weighted binary cross-entropy loss $L^{MF}$ combined with a sigmoid layer in the moral foundation classifier:
\[w_{c} = \frac{\#\ negative\ examples\ in\ c}{\#\ positive\ examples\ in\ c}\]
\[L^{MF} = \frac{1}{N} \sum_{n=1}^{N} -w_{c} \cdot y^{c} \cdot log \sigma(x_{n}^{c}) + (1-y^{c})\cdot log(1-\sigma(x_{n}^{c}))\]
where \(c\) is the class a data example belongs to (eg. care), \(w_c\) is the weight for this class, \(n\) is the data example index, \(x_{n}^{c}\) is the final-layer embedding of the \(n^{th}\) example which belongs to the class \(c\), \(\sigma()\) represents the sigmoid function, and \(y^{c}\) is the label.

\paragraph{Domain Classifier}

This is the adversary in the model. It distinguishes which domain an input example comes from and pushes the BERT encoder to produce domain-invariant embeddings that can fool the domain classifier itself. The domain classifier has the same feed-forward network structure as the moral foundation classifier, but a different loss \(L^{D}\) - a cross-entropy loss with a softmax layer. This loss is to be \textit{maximized} during the adversarial training (our goal is to fool this domain classifier), which is achieved by connecting the domain classifier to the other parts of the model with a gradient reversal layer \cite{ganin2016domain}.

\paragraph{Reconstruction Module}

To compete with the adversarial domain classifier, the BERT encoder wants to wipe out as much domain-specific information as possible. However, if not controlled well, the BERT encoder could be overly corrupted, and too much information would be lost in the representations learned. Therefore, we add a reconstruction module to ensure the BERT encoder can still generate representations close to the original contextual BERT embeddings. The reconstruction module has a linear layer followed by a tanh activation. Its loss $L^{rec}$ is calculated as the mean squared error between the reconstructed embeddings and the original embeddings generated by BERT with no domain adversarial training:
\[L^{rec} = || tanh(Wx+b) - tanh(x_{orig}) ||^2\]
where \(x\) is the embeddings generated by the current model, $W$ and $b$ are the weight and bias for a linear transformation on $x$, and \(x_{orig}\) is the embeddings generated by BERT model with no domain adversarial training.

\hfill

The overall loss function of the whole model is:
\[L = \lambda^{rec} \cdot L^{rec} + \lambda^{trans} \cdot L^{trans} + L^{MF} - \lambda^{D} \cdot L^{D}\]
\noindent where \(\lambda^{rec}\), \(\lambda^{trans}\) and \(\lambda^{D}\) are hyper-parameters to be tuned while training. 
The first three terms in this loss are \textit{minimized} with respect to the model parameters, whereas the last loss term for the domain classifier adversary is \textit{maximized} with respect to its parameters. This is to achieve the objective of fooling the domain classifier and generating domain-invariant embeddings. During training, the maximization of the domain classifier loss term is achieved by using a gradient reversal layer \cite{ganin2016domain}.

\section{Experiments}

\subsection{Data}

We evaluated the performance of the proposed model on labeled moral foundations data in the out-of-domain setting. 

\paragraph{Moral Foundations Twitter Corpus (MFTC)}
This dataset \cite{Hoover2020moral} includes 35K English tweets, and is labeled by trained human annotators with moral foundation labels in 11 classes (eg ``care,'' ``harm,'' ``fairness,'' etc), including a ``non-moral'' class. The tweets cover six different topics: (1) Black lives matter (BLM) civil protests, (2) All Lives Matter (ALM), (3) the Baltimore protest against the death of Freddie Grey, (4) the 2016 USA Presidential election, (5) Hurricane Sandy and (6) hate speech. We follow the standard practice and aggregate the annotations from multiple annotators using majority vote as the true label. 
We discard tweets that don't reach a majority vote.

\paragraph{Covid}
This dataset \cite{rojecki2021moral} contains 2,648 tweets related to the COVID19 pandemic. These tweets were manually annotated with 10 moral foundation classes.

\paragraph{Congress}
This dataset is a  collection of 2,050  English-language tweets by members of US Congress~\cite{johnson-goldwasser-2018-classification}. The dataset covers six different political topics: (1) abortion, (2) the Affordable Care Act, (3) guns, (4) immigration, (5) LGBTQ rights, and (6) terrorism.

\paragraph{The extended Moral Foundations Dictionary (eMFD)}
This dataset differs in terms of source of text and the annotation process. It contains 2995 English news articles on a variety of topics. The labeling task was  crowd-sourced to 557 annotators, who were asked to mark all text segments  expressing a given moral foundation in 15 randomly selected news articles~\cite{hopp2021extended}. After processing, this dataset contains 36K labeled examples.

\hfill

\begin{table}[h]
\centering
\begin{tabular}{lll}
\hline
    \textbf{Data} & \textbf{Type} & \textbf{Size} \\
    \hline
    MFTC & tweets & 18991 \\
    Covid & tweets & 2430 \\
    Congress & tweets & 1849 \\
    eMFD & news articles & 35985 \\
    \hline
\end{tabular}
\caption{An overview of the datasets}
\label{dataset}
\end{table}

\begin{figure*}[!t]
\centering
\includegraphics[width=\textwidth]{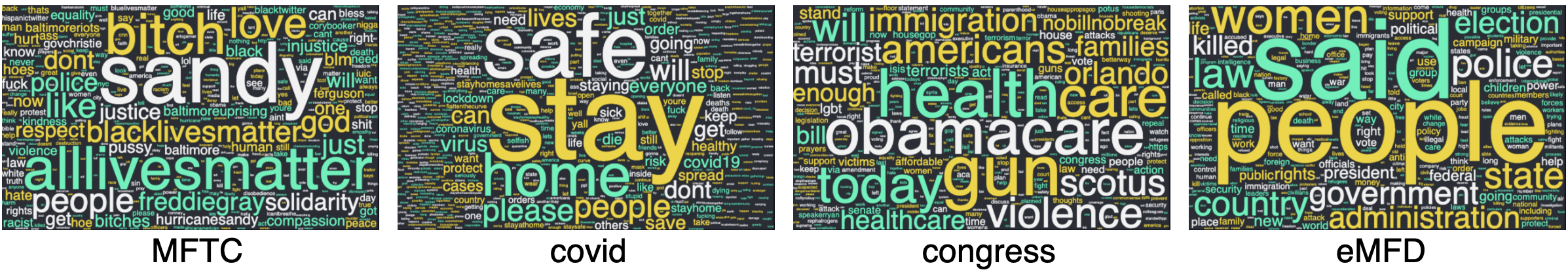}
\caption{Word clouds for each dataset.}
\label{fig_word_clouds}
\end{figure*}

Table \ref{dataset} shows the sizes of the four datasets after preprocessing. Figure \ref{fig_word_clouds} demonstrates the distinct topics discussed in these data by word clouds. We process the texts in all datasets by removing URLs, replacing all mentions with ``@user'', removing hashtags, replacing emojis with their text descriptions\footnote{\url{https://pypi.org/project/emoji/}}, and removing all non-ASCII characters. We took the maximum sequence length to be 50 when training with BERT and DAMF, because the majority of the datasets are tweets or segments in news articles with short lengths. Each data example has binary labels in 10 moral classes, and a non-moral example would have ``False'' for all 10 classes.

We test different models on four datasets separately. 
For each test dataset, we experiment with training on a single dataset or combinations of multiple labeled datasets.

\subsection{Model Training}

We compare our proposed model \textbf{DAMF} with three baseline models, (1) Distributed Dictionary Representations (DDR) \cite{8b2871f503a14011ae81e6ab1664a638}, (2) BERT \cite{devlin2019bert} and (3) BERT improved with the Lightweight Adversarial Filtering (AFLite) method \cite{le2020adversarial}. 

To evaluate the models, we use F1 score weighted by the number of true examples for each class. Each experiment is repeated with 5 random seeds to calculate the mean and the standard deviation.

\paragraph{Distributed Dictionary Representations (DDR)}
DDR \cite{8b2871f503a14011ae81e6ab1664a638} is an unsupervised lexicon-based method built on the Moral Foundations Dictionary (MFD), and it is a classic method widely used in prior works \cite{garten2016morality,Hoover2020moral,trager2022moral}. It represents each moral foundation class by the centroid of the embedding vectors of all words in this category of MFD, and then calculates the cosine similarity between an input text (as the average of its word embeddings) with each of the moral foundation class embedding. Here we use word2vec word embeddings \cite{mikolov2013efficient}. The moral foundation class label is predicted as the one with the largest similarity score to the input text.

\paragraph{BERT}
BERT is used in prior works to detect morality \cite{roy-goldwasser-2021-analysis} and is \textit{the current state-of-the-art method for morality prediction} \cite{trager2022moral}. As a supervised model, BERT can perform representation learning with respect to specific tasks or datasets and thus improves the performance. Therefore, finetuning BERT offers us a strong baseline. 

We finetune BERT with a linear prediction layer attached. To mitigate over-fitting, we implement a dropout layer with a rate of 0.3. For multi-label prediction, we use the binary cross-entropy loss with sigmoid function. After the naive dataset merge is done, we split the training dataset into 80-20 train-validation sets, and we predict on separate test data. We train different datasets with 20 epochs and stop early when the model achieves the best validation F1 score. The batch size is 64. We use the Adam optimizer with an initial learning rate of 5e-5, and we use a scheduler
\(lr = lr_{init}/((1 +\alpha \cdot p)^\beta)\), where \(p = \frac{current\ epoch}{total\ epoch}\) and we set \(\alpha=10\) and \(\beta=0.25\).

\paragraph{BERT+AFLite}

To have an even stronger baseline incorporating domain adaptation techniques, we select BERT combined with the Lightweight Adversarial Filtering (BERT+AFLite). AFLite filters out the data points that are too easy to be predicted by weak classifiers and minimizes the ability of a model to exploit spurious correlations, thus improves model's out-of-domain performance \cite{sakaguchi2021winogrande, le2020adversarial}. We use a single linear layer of neural network as the weak classifier, iteratively train and test on different random partitions of the data, and discard texts if the linear classifier can predict their moral foundation using vanilla BERT embedding. We perform iterative filtering until the data size is shrunk to 50\% or until the data size changes from the last round by less than 2\%, whichever is achieved first. AFLite filtered out \textit{MFTC} by 50\%, the \textit{covid} data by 10\%, the \textit{congress} data by 15\%, and \textit{eMFD} by 28\%. Then, we train a BERT model with the filtered data using the same procedure described in the last section.

\paragraph{DAMF}
We train the DAMF model in a semi-supervised fashion. That is, in addition to the usual labeled training data, we also use unlabeled data from the target domain in each training data batch. This way, the domain adapting module in our model can align the feature distributions of different datasets including the target data. The data from different domains in each batch are balanced in size. The morality labeled training data passes through both the moral foundation classifier and the domain classifier, and the unlabeled data in the target domain only passes through the domain classifier. We then evaluate on a separate hold-out test dataset.

We train different datasets with 60 epochs in total. During the first 15 epochs, the model is trained without the domain adversarial module activated. This is for the model to better learn the morality labels first. The batch size is 64. We use the Adam optimizer. The learning rate and the scheduler are the same as for BERT model. 

When training with a domain adversary, it is challenging to control its power. If too powerful, it easily wipes out useful information in the feature embeddings and leads to bad moral foundation predictions. We control the hyperparamters \(\lambda^{rec}\), \(\lambda^{trans}\) and \(\lambda^{D}\). Following Ganin et al. (\citeyear{ganin2015unsupervised}) and Allaway et al. (\citeyear{allaway2021adversarial}), we control \(\lambda^{D}\) indirectly by a parameter \(\gamma\):
\[\lambda^{D}=2/(1+e^{-\gamma \cdot p})-1\]
where \(p = \frac{current\ epoch - epochs\ trained\ w/o\ adversary}{total\ epochs}\). Thus, we perform hyperparameter search on \(\lambda^{rec}\) in \([0, 0.1, 0.5, 1]\), \(\lambda^{trans}\) in \([no\ trans, 0.01, 0.1, 1, 10]\) and \(\gamma\) in \([0.1, 1, 10]\) (see Appendix for hyperparmeters we used).

\section{Results}

\subsection{Heterogeneity in Feature and Label spaces}

\begin{figure}[h]
\centering
\includegraphics[width=0.41\textwidth]{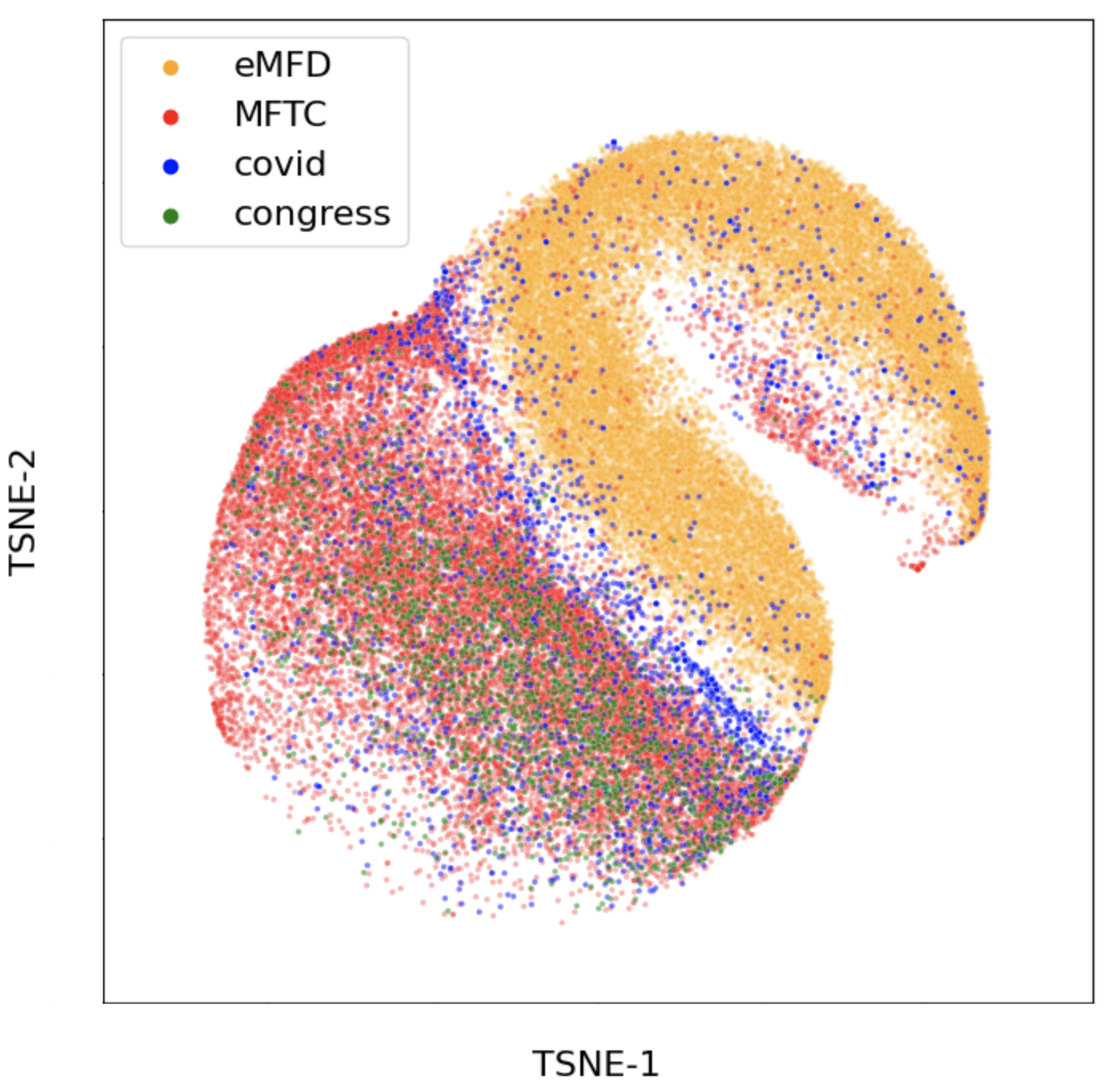}
\caption{Feature distributions of different datasets, visualized with TSNE. For each text example (dots), we compute its embeddings using a vanilla BERT model without training, and then compute the first and second TSNE components.}
\label{fig_feature_distr}
\end{figure}

We first examine the heterogeneity of different datasets and show the two critical aspects that need to be considered 
during multi-domain learning: 
how datasets differ in the feature space and how they differ in label space. In the feature space (Fig.~\ref{fig_feature_distr}), different datasets have different distributions of embeddings. For example, the \textit{congress} dataset, which focuses on messages on US political issues, only takes a subset in the TSNE space compared to the \textit{MFTC} dataset, which includes messages posted on social media on six various topics. 
Another example \textit{eMFD} contains text of news articles and is also annotated by a crowd. It is very different from other datasets, which have been annotated by a few experts. In Fig.~\ref{fig_feature_distr}, \textit{eMFD} shows a more distinct distribution compared to all other Twitter-based datasets.  We therefore expect that a model trained on \textit{eMFD} would not generalize well to the other datasets (Table \ref{tab_results} row 3, row 9, row 16, BERT model performance).

\begin{figure}[h]
\centering
\includegraphics[width=0.42\textwidth]{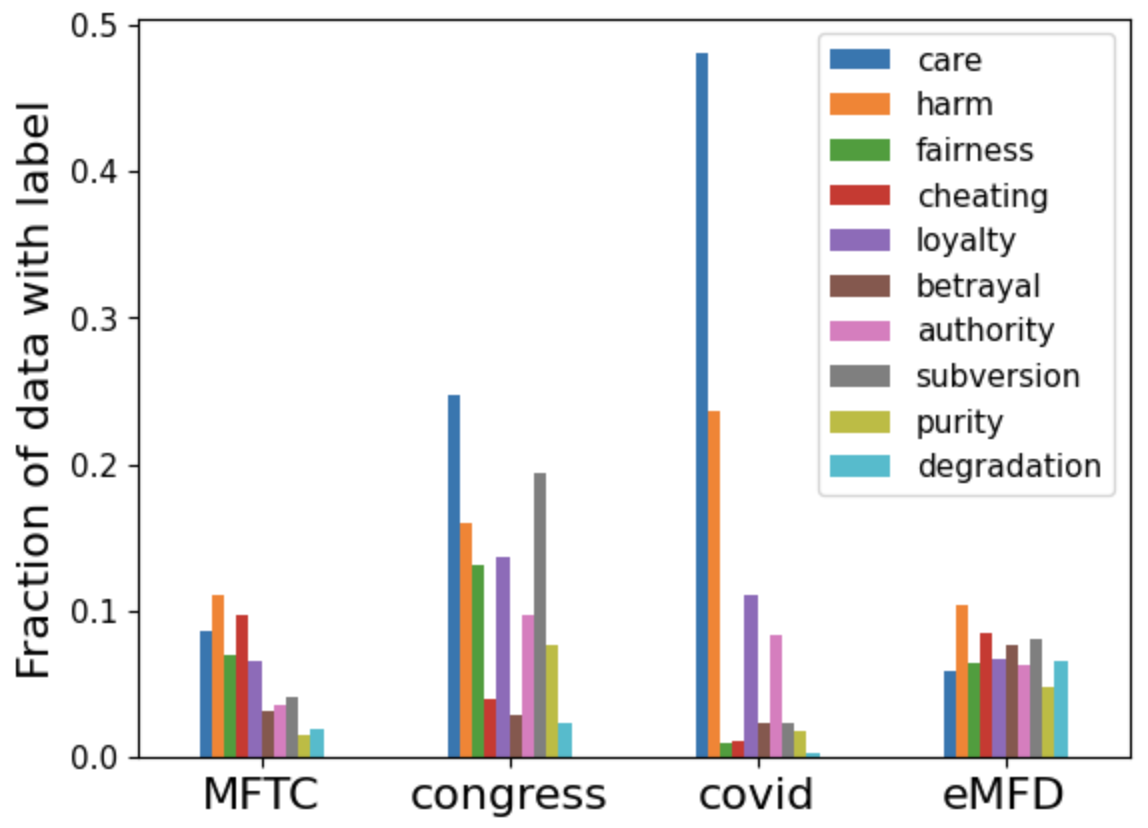}
\caption{Label distributions of different datasets. The fraction of text examples with each label differs significantly among the datasets.}
\label{fig_label_distr}
\end{figure}


In addition to features, the label distributions among heterogeneous datasets can also differ dramatically (Fig.~\ref{fig_label_distr}). This is because people express very different moral concerns on different topics on different platforms or settings. For example, the larger and more general \textit{MFTC} data has a broad distribution of texts from all moral classes, but the \textit{covid} dataset is heavily biased toward the \textit{care/harm} foundation. People expressed more support and care when talking about pandemic-related topics such as ``stay at home'' (Fig. \ref{fig_word_clouds}). This imbalance among classes can fundamentally affect multi-domain learning. For example, a model trained on the \textit{covid} dataset learns very little about fairness. When a test data such as \textit{MFTC} has many protest-related posts expressing fairness (Fig. \ref{fig_word_clouds} and Fig. \ref{fig_label_distr}), the model doesn't predict this label accurately (Table \ref{tab_results} row 1, BERT model performance). 

\subsection{Outstanding Performance of DAMF}


\begin{table*}[t]
\centering
\begin{tabular}{c|cc|ccccc}
\hline
    & \textbf{Training Data} & \textbf{Test Data} & \textbf{DDR} & \textbf{BERT} & \textbf{BERT+AFLite} & \thead{\textbf{DAMF}\\-\textbf{weightedL}} & \thead{\textbf{DAMF}\\\textbf{full model}} \\
    \hline
    1 & covid & MFTC & \textbf{0.38} & $0.20 \pm 0.01$ & $0.20 \pm 0.02$  & $0.23 \pm 0.04$ & $0.26 \pm 0.03$\\
    2 & congress & MFTC & \textbf{0.38} & $ 0.31 \pm 0.02$ & $0.30 \pm 0.01$ & $0.33 \pm 0.01$ & $0.35 \pm 0.02$ \\
    3 & eMFD & MFTC & \textbf{0.38} & $0.22 \pm 0.03$ & $0.20 \pm 0.02$ & $0.30 \pm 0.04$ & $0.34 \pm 0.02$ \\
    4 & covid + eMFD & MFTC & \textbf{0.38} & $0.24 \pm 0.02$ & $0.28 \pm 0.02$ & $0.27 \pm 0.03$ & $0.34 \pm 0.02$ \\
    5 & covid + congress & MFTC & 0.38 & $0.35 \pm 0.01 $ & $0.34 \pm 0.02$ & $ 0.38 \pm 0.01 $ & $ \textbf{0.42} \pm \textbf{0.01} $\\
    6 & eMFD + congress & MFTC & 0.38 & $0.34 \pm 0.02 $ & $0.29 \pm 0.02$ & $ 0.36 \pm 0.02 $ & $ \textbf{0.39} \pm \textbf{0.02} $\\
    7 & covid + congress + eMFD & MFTC & 0.38 & $0.31 \pm 0.01$ & $0.31 \pm 0.02$ & $0.33 \pm 0.03$ & $\textbf{0.43} \pm \textbf{0.01}$ \\
    \hline
    8 & covid & congress & 0.23 & $0.10 \pm 0.05 $ & $0.26 \pm 0.01$ & $ 0.12 \pm 0.02 $ & $ \textbf{0.35} \pm \textbf{0.02} $ \\
    9 & eMFD & congress & 0.23 & $0.14 \pm 0.02$ & $0.13 \pm 0.02$ & $0.21 \pm 0.05$ & $\textbf{0.33} \pm \textbf{0.01}$ \\
    10 & MFTC & congress & 0.23 & $0.30 \pm 0.02$ & $0.35 \pm 0.01$ & $0.30 \pm 0.03$ & $\textbf{0.38} \pm \textbf{0.01}$ \\
    11 & covid + eMFD & congress & 0.23 & $0.26 \pm 0.03 $ & $0.21 \pm 0.06$ & $ 0.28 \pm 0.02$ & $\textbf{0.35} \pm \textbf{0.01}$ \\
    12 & covid + MFTC & congress & 0.23 & $0.31 \pm 0.02 $ & $0.35 \pm 0.01$ & $ 0.34 \pm 0.02 $ & $ \textbf{0.40} \pm \textbf{0.03} $\\
    13 & eMFD + MFTC & congress & 0.23 & $0.28 \pm 0.05$ & $0.34 \pm 0.01$ & $0.30 \pm 0.03$ & $\textbf{0.37} \pm \textbf{0.02}$ \\
    14 & covid + eMFD + MFTC & congress & 0.23 & $0.30 \pm 0.01$ & $0.32 \pm 0.01$ & $0.33 \pm 0.02$ & $\textbf{0.38} \pm \textbf{0.02}$ \\
    \hline
    15 & congress & covid & 0.43 & $0.57 \pm 0.01$ & $\textbf{0.59} \pm \textbf{0.02}$ & $0.58 \pm 0.01$ & $\textbf{0.59} \pm \textbf{0.02}$ \\
    16 & eMFD & covid & 0.43 & $0.32 \pm 0.03$ & $0.23 \pm 0.12$ & $0.40 \pm 0.05$ & $\textbf{0.51} \pm \textbf{0.02}$ \\
    17 & MFTC & covid & 0.43 & $0.49 \pm 0.02$ & $\textbf{0.61} \pm \textbf{0.01}$ & $0.52 \pm 0.01$ & $0.56 \pm 0.03$ \\
    18 & congress + eMFD & covid & 0.43 & $0.47 \pm 0.02$ & $0.30 \pm 0.02$ & $0.51 \pm 0.02$ & $\textbf{0.57} \pm \textbf{0.01}$ \\
    19 & congress + MFTC & covid & 0.43 & $0.51 \pm 0.01$ & $\textbf{0.61} \pm \textbf{0.01}$ & $0.56 \pm 0.02$ & $\textbf{0.61} \pm \textbf{0.03}$ \\
    20 & eMFD + MFTC & covid & 0.43 & $0.45 \pm 0.04$ & $\textbf{0.57} \pm \textbf{0.03}$ & $0.47 \pm 0.05$ & $0.56 \pm 0.02$ \\
    21 & congress + eMFD + MFTC & covid & 0.43 & $0.49 \pm 0.01$ & $0.58 \pm 0.02$ & $0.52 \pm 0.01$ & $\textbf{0.61} \pm \textbf{0.02}$ \\
    \hline
    22 & covid & eMFD & 0.17 & $0.12 \pm 0.01$ & $0.13 \pm 0.01$ & $0.13 \pm 0.02$ & $\textbf{0.18} \pm \textbf{0.01}$ \\
    23 & congress & eMFD & 0.17 & $0.16 \pm 0.01$ & $0.17 \pm 0.01$ & $0.17 \pm 0.01$ & $\textbf{0.20} \pm \textbf{0.01}$ \\
    24 & MFTC & eMFD & 0.17 & $0.13 \pm 0.01$ & $0.17 \pm 0.01$ & $0.16 \pm 0.01$ & $\textbf{0.18} \pm \textbf{0.01}$ \\
    25 & covid + congress & eMFD & 0.17 & $0.14 \pm 0.01$ & $0.16 \pm 0.01$ & $0.15 \pm 0.01$ & $\textbf{0.20} \pm \textbf{0.01}$ \\
    26 & covid + MFTC & eMFD & 0.17 & $0.14 \pm 0.01$ & $0.17 \pm 0.01$ & $0.15 \pm 0.01$ & $\textbf{0.19} \pm \textbf{0.01}$ \\
    27 & congress + MFTC & eMFD & 0.17 & $0.14 \pm 0.01$ & $0.17 \pm 0.01$ & $0.16 \pm 0.01$ & $\textbf{0.19} \pm \textbf{0.01}$ \\
    28 & covid + congress + MFTC & eMFD & 0.17 & $0.15 \pm 0.01$ & $0.17 \pm 0.01$ & $0.16 \pm 0.01$ & $\textbf{0.21} \pm \textbf{0.01}$ \\
    \hline
\end{tabular}
\caption{Comparison between baseline models and \textbf{DAMF}. The numbers are F1 scores weighted by the number of true instances for each class. Each experiment is repeated with 5 random seeds to calculate the standard deviation. DAMF outperforms the baselines in most of the scenarios. The ablation study shows that both the domain adversarial module and the weighted loss function \(L^{MF}\) in moral foundation classifier play important roles. }
\label{tab_results}
\end{table*}

We demonstrate how fusing these heterogeneous datasets with \textbf{DAMF} helps to tackle both feature and label shifts, and improves the model performance and generalizability. Table \ref{tab_results} shows the outstanding performance of \textbf{DAMF}. In most of the settings, it outperforms other baselines. 

First, we compare \textbf{DAMF} with the widely used unsupervised method, DDR. It does not require training and thus is less susceptible to the training/test data shift problem. From Table \ref{tab_results} row 1 to 4, we see that DDR is not affected when training data are small and limited, which is an advantage when predicting on the larger general test data \textit{MFTC}. On the other hand, the supervised methods BERT and \textbf{DAMF} have more difficulty when they are trained on smaller datasets. Nevertheless, as more data with annotations becomes available, supervised models gain more benefits. By merging these datasets and using a good model to improve generalizability, the full \textbf{DAMF} model outperforms DDR on all datasets.

Next, we compare \textbf{DAMF} with the current state-of-the-art morality prediction model, BERT. It is widely agreed that simply finetuning BERT can lead to overfitting on the training data and worsen its performance. Our results have shown that \textbf{DAMF} with domain adaptation ability fully outperforms the vanilla BERT, especially when the training and test data have more distinct feature and/or label distributions. One example is when training on \textit{eMFD} and testing on \textit{congress} (Table \ref{tab_results} row 9), where these two datasets' feature and label distributions are both distinct, \textbf{DAMF} outperforms BERT by 135\% in F1-score.

We then discuss how \textbf{DAMF} compares to the strongest baseline, BERT+AFLite. On top of vanilla BERT, this method incorporates domain adaptation technique, filtering out overly easy training data points which may contain spurious correlation, and thus improves out-of-domain prediction performance. However, our results show that this method is not robust for all datasets. The \textit{covid} dataset has a label distribution very concentrated on care and harm, and thus is easier to predict. In this case, BERT+AFLite has a good performance comparable to \textbf{DAMF} (Table \ref{tab_results} row 15, 17, 19 and 20). Nevertheless, when applied on \textit{eMFD} with noisier texts and label distributions, AFLite even hurts the performance (e.g. BERT+AFLite has worse F1 than BERT in Table \ref{tab_results} row 6 and 11), possibly because AFLite mistakenly filters out relevant but noisy data. 

In addition, we also observe from Table \ref{tab_results} that combining datasets for training generally improves BERT, BERT+AFLite and \textbf{DAMF}'s performances. Nevertheless, it is important to consider what datasets to merge. For example, \textit{eMFD} has a more distinct feature distribution from others. When we merge \textit{eMFD} and other dataset for training, it doesn't always improve the performance (e.g. comparing Table \ref{tab_results} row 5 and 7, row 12 and 14). By merging the suitable datasets with different feature and label distributions, we can facilitate the out-of-domain prediction by having the training data cover a wider feature distribution, and also by compensating each other to mitigate the label imbalance.

These above results show \textbf{DAMF}'s ability to handle feature and label shifts and to better adapt when learning multiple heterogeneous datasets. Table \ref{tab_examples} shows some examples where \textbf{DAMF} successfully detected the correct moral foundations when other models fail. 
Row 3 and 4 show that BERT+AFLite does not always give robust predictions. By outperforming strong baselines like BERT and BERT+AFLite in most experiments, \textbf{DAMF} is shown to achieve state-of-the-art performance for morality detection in out-of-domain testing.

\begin{table*}[ht]
\centering
\begin{tabular}{l|l|llll}
\hline
    & & \multicolumn{4}{c}{\textbf{predictions}}  \\
    \hline
    \textbf{text} & \textbf{ground truth} & \textbf{DDR} & \textbf{BERT} & \textbf{BERT+AFLite} & \textbf{DAMF} \\
    \hline
    \specialcell{Do u know how selfish you're being to\\put other peoples lifes at risk too????} & betrayal & degradation & care & care & betrayal,harm \\
    \hline
    \specialcell{doing what they are supposed to be\\doing \#socialdistancing} & authority & fairness & subversion & subversion & authority \\
    \hline
    Stay home stay safe & care & care & non-moral & fairness & care \\
    \hline
    stay safe and stay home & care & care & non-moral & cheating & care\\
    \hline
\end{tabular}
\caption{Examples of \textbf{DAMF} successfully detected the correct moral foundations when other models fail. Row 3 and 4 also show that BERT+AFLite does not give robust predictions.}
\label{tab_examples}
\end{table*}

\subsection{Ablation Study} 

\begin{figure}[h]
\centering
\includegraphics[width=0.48\textwidth]{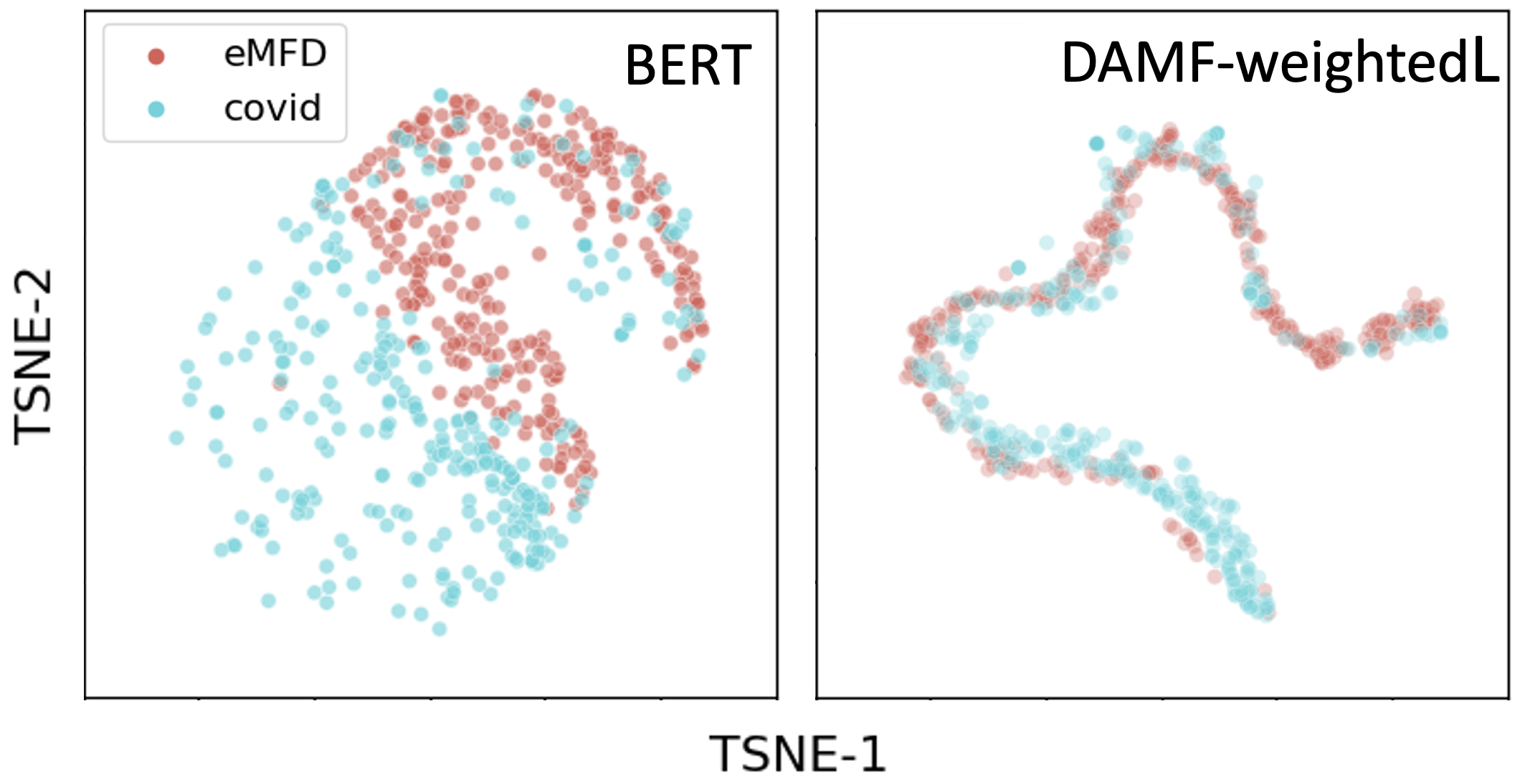}
\caption{The feature embedding distributions of random samples of \textit{eMFD} and \textit{covid} datasets learned by vanilla BERT (left) and DAMF-weightedL (right), visualized with TSNE. DAMF-weightedL successfully aligns the feature distribution of these two distinct datasets.}
\label{fig_compare_tsne}
\end{figure}

To verify that both the domain adversarial module and the weighted loss function help improve model generalizability, we perform an ablation study (the last two columns in Table \ref{tab_results}). The DAMF-weightedL model uses the domain adversarial module and only a regular binary cross-entropy loss function. Figure \ref{fig_compare_tsne} shows an example where the DAMF-weightedL model successfully aligns the distinct feature distributions between \textit{eMFD} and \textit{covid}. Furthermore, from Table \ref{tab_results} we see DAMF-weightedL already outperforms BERT model, showing the benefit of using the domain adversarial module. We also notice that the domain adversarial module is more effective when adapting between datasets with very distinct feature distributions (e.g. from \textit{eMFD} to \textit{congress} in row 9, DAMF-weightedL model outperforms BERT by 50\% in F1). The domain adversarial module also tends to be more effective when adapting from smaller limited training datasets to larger general datasets (e.g. from \textit{covid} to \textit{MFTC} in row 1, DAMF-weightedL model outperforms BERT by 15\% in F1). It is less effective when adapting from an already general training dataset to a limited test dataset, 
as the test data embeddings might already align with training data embeddings. 

Next, we look at the performance of the full \textbf{DAMF} model which includes both the domain adversarial module and the weighted loss function. In all cases, the full model outperforms the DAMF-weightedL model. This says that the weighted loss function also plays an important role. This is especially true when the training data is \textit{covid}. \textit{covid} has the most imbalanced label distribution that is different from other datasets' label distributions, thus the weighted loss function gives a boost in performance. For example, when adapting from \textit{covid} to \textit{congress} (Table \ref{tab_results} row 8), the full \textbf{DAMF} outperforms the DAMF-weightedL model by 190\%, and Table \ref{tab_weighted_loss} shows the improved and more balanced per-class F1-scores by \textbf{DAMF}. 

\begin{table*}[ht]
\centering
\begin{tabular}{l|l|lll|lll}
\hline
    & & \multicolumn{3}{c|}{\textbf{DAMF}-\textbf{weightedL}} & \multicolumn{3}{c}{\textbf{DAMF}} \\
    \hline
    & \textbf{support} & \textbf{precision} & \textbf{recall} & \textbf{F1} & \textbf{precision} & \textbf{recall} & \textbf{F1} \\
    \hline
    care & 100 & 0.50 & 0.54 & 0.52 & 0.39 & 0.69 & 0.50 \\
    harm & 62 & 0.47 & 0.42 & 0.44 & 0.35 & 0.56 & 0.43 \\
    fairness & 45 & 0.00 & 0.00 & 0.00 & 0.06 & 0.11 & 0.08 \\
    cheating & 17 & 0.00 & 0.00 & 0.00 & 0.09 & 0.41 & 0.15 \\
    loyalty & 51 & 0.30 & 0.53 & 0.38 & 0.20 & 0.90 & 0.33 \\
    betrayal & 12 & 0.00 & 0.00 & 0.00 & 0.06 & 0.50 & 0.11 \\
    authority & 36 & 0.13 & 0.19 & 0.16 & 0.12 & 0.97 & 0.21 \\
    subversion & 73 & 0.00 & 0.00 & 0.00 & 0.29 & 0.68 & 0.41 \\
    purity & 31 & 0.00 & 0.00 & 0.00 & 0.13 & 0.97 & 0.22 \\
    degradation & 10 & 0.00 & 0.00 & 0.00 & 0.03 & 0.70 & 0.06 \\
    \hline
\end{tabular}
\caption{A comparison of DAMF-weightedL and \textbf{DAMF} on per-class performance. \textbf{DAMF} with the weighted loss has a more balanced performance among different label classes. Both models are trained on \textit{covid} and tested on a sample of \textit{congress} of size 370. }
\label{tab_weighted_loss}
\end{table*}

\section{Conclusions}

We have proposed a data fusion framework, \textbf{DAMF}, for training moral foundation classifiers on heterogeneous datasets. We demonstrate the benefit of merging suitable datasets and show that \textbf{DAMF} outperforms three different baselines in various experimental settings. We also show that the domain adversarial module and the weighted loss function help align the distributions of data in the feature space and the label space, respectively. 

There are limitations in our work that requires future improvements. By merging heterogeneous datasets using \textbf{DAMF}, we hope to reduce the differences between datasets due to varying procedures for data collection, annotation, topic selection, etc. One limitation is that we do not consider the complexities of various topics existing in single datasets like \textit{MFTC}. One of our future works is to disaggregate the large datasets into different topics, and apply domain adaptation methods on separate topics. In addition, we could even disaggregate datasets by considering each annotator as a single domain, and target the differences from the annotation processes~\cite{zhang2021crowdsourcing}.

To mitigate label shift between training and test data, the weighted loss function is a simple method. In the future, we plan to explore more sophisticated methods for domain adaptation targeting label shift, for example, estimating the relative class weights between domains and sample re-weighting as in the work of Tachet et al. (\citeyear{tachet2020domain}), or using additional loss functions to account for conflicting and co-occurring labels to achieve better generalization performance with respect to label shift \cite{kim2022learning}.

\section{Ethical Considerations}
Morality is relatively personal and subjective. It can vary based on different individuals, backgrounds or cultures. In this study, we focus on English language, American culture, and the morality expressed in news articles or on social media platforms.  Unfortunately, in the datasets or the pre-trained models that we have used, it is possible that biases exist with respect to gender, race, or other factors.

\section{Acknowledgements}
This project was funded in part by DARPA under contract HR001121C0168.

\bibliography{aaai23}

\section{Appendix}

\begin{table*}[ht]
\centering
\begin{tabular}{c|cc|ccc}
\hline
    & \textbf{Training Data} & \textbf{Test Data} & \textbf{\(\lambda^{trans}\)} & \textbf{\(\lambda^{rec}\)} & \textbf{\(\gamma\)} \\
    \hline
    1 & covid & MFTC & 0.01 & 1 & 1\\
    2 & congress & MFTC & 0.1 & 0.1 & 10 \\
    3 & eMFD & MFTC & no trans & 0 & 0.1 \\
    4 & covid + eMFD & MFTC & 1 & 0 & 0.1 \\
    5 & covid + congress & MFTC & 1 & 1 & 1 \\
    6 & eMFD + congress & MFTC & 0.1 & 1 & 10 \\
    7 & covid + congress + eMFD & MFTC & no trans & 0 & 0.1 \\
    \hline
    8 & covid & congress & 0.1 & 0.1 & 1 \\
    9 & eMFD & congress & 10 & 0.1 & 1 \\
    10 & MFTC & congress & no trans & 1 & 1 \\
    11 & covid + eMFD & congress & 0.1 & 0.1 & 1 \\
    12 & covid + MFTC & congress & no trans & 0.1 & 1 \\
    13 & eMFD + MFTC & congress & no trans & 0 & 0.1 \\
    14 & covid + eMFD + MFTC & congress & no trans & 0 & 0.1 \\
    \hline
    15 & congress & covid & 10 & 0.5 & 10 \\
    16 & eMFD & covid & 0.1 & 0.5 & 1 \\
    17 & MFTC & covid & no trans & 0 & 0.1 \\
    18 & congress + eMFD & covid & 0.01 & 1 & 10 \\
    19 & congress + MFTC & covid & 10 & 0 & 0.1 \\
    20 & eMFD + MFTC & covid & no trans & 0.1 & 1 \\
    21 & congress + eMFD + MFTC & covid & no trans & 0 & 0.1 \\
    \hline
    22 & covid & eMFD & 0.01 & 0.1 & 1 \\
    23 & congress & eMFD & 0.01 & 0 & 1 \\
    24 & MFTC & eMFD & no trans & 0 & 0.1 \\
    25 & covid + congress & eMFD & 0.1 & 1 & 1 \\
    26 & covid + MFTC & eMFD & no trans & 0 & 1 \\
    27 & congress + MFTC & eMFD & no trans & 0 & 0.1 \\
    28 & covid + congress + MFTC & eMFD & 0.1 & 0 & 0.1 \\
    \hline
\end{tabular}
\caption{Hyperparameters used for DAMF-weightedL model and the full DAMF model to produce results shown in Table \ref{tab_results}.}
\label{hp}
\end{table*}

\end{document}